\documentclass[10pt,twocolumn,letterpaper]{article}

\usepackage[pagenumbers]{cvpr} % To force page numbers, e.g. for an arXiv version

\usepackage{graphicx}%
\usepackage{amsmath,amssymb,amsfonts}%
\usepackage{amsthm}%
\usepackage{mathrsfs}%
\usepackage[title]{appendix}%
\usepackage{textcomp}%
\usepackage{manyfoot}%
\usepackage[export]{adjustbox}
\usepackage{listings}%
\usepackage{booktabs}%
\usepackage{bm}
\usepackage{setspace}
\usepackage{makecell}
\usepackage{bm}
\usepackage{stfloats}
\usepackage{multirow}
\usepackage{cuted}
\usepackage{marvosym, ifsym} % for corresponding author letter symbol

\definecolor{cvprblue}{rgb}{0.21,0.49,0.74}
\usepackage[pagebackref,breaklinks,colorlinks,allcolors=cvprblue]{hyperref}

\title{ProEdit: Inversion-based Editing From Prompts Done Right}

\author{Zhi Ouyang$^{1,*}$
\quad
Dian Zheng$^{2,*}$
\quad
Xiao-Ming Wu$^{3}$
\quad
Jian-Jian Jiang$^{1}$\\
\quad
Kun-Yu Lin$^{4}$
\quad
Jingke Meng\textsuperscript{1,\Letter}
\quad
Wei-Shi Zheng\textsuperscript{1,5,\Letter}\\\\
$^{1}$Sun Yat-sen University
\quad
$^{2}$CUHK MMLab\\
$^{3}$College of Computing and Data Science, Nanyang Technological University\\
$^{4}$The University of Hong Kong\\
$^{5}$Key Laboratory of Machine Intelligence and Advanced Computing, Ministry of Education, China\\
\tt\normalsize\color{Magenta}\url{https://isee-laboratory.github.io/ProEdit/} \\\\%[-3pt]
}

\begin{document}
\maketitle
\captionsetup{hypcap=false} % 关闭 hypcap 以避免警告
\begin{strip}\centering
\vspace{-1.4cm}
\includegraphics[width=1\linewidth]{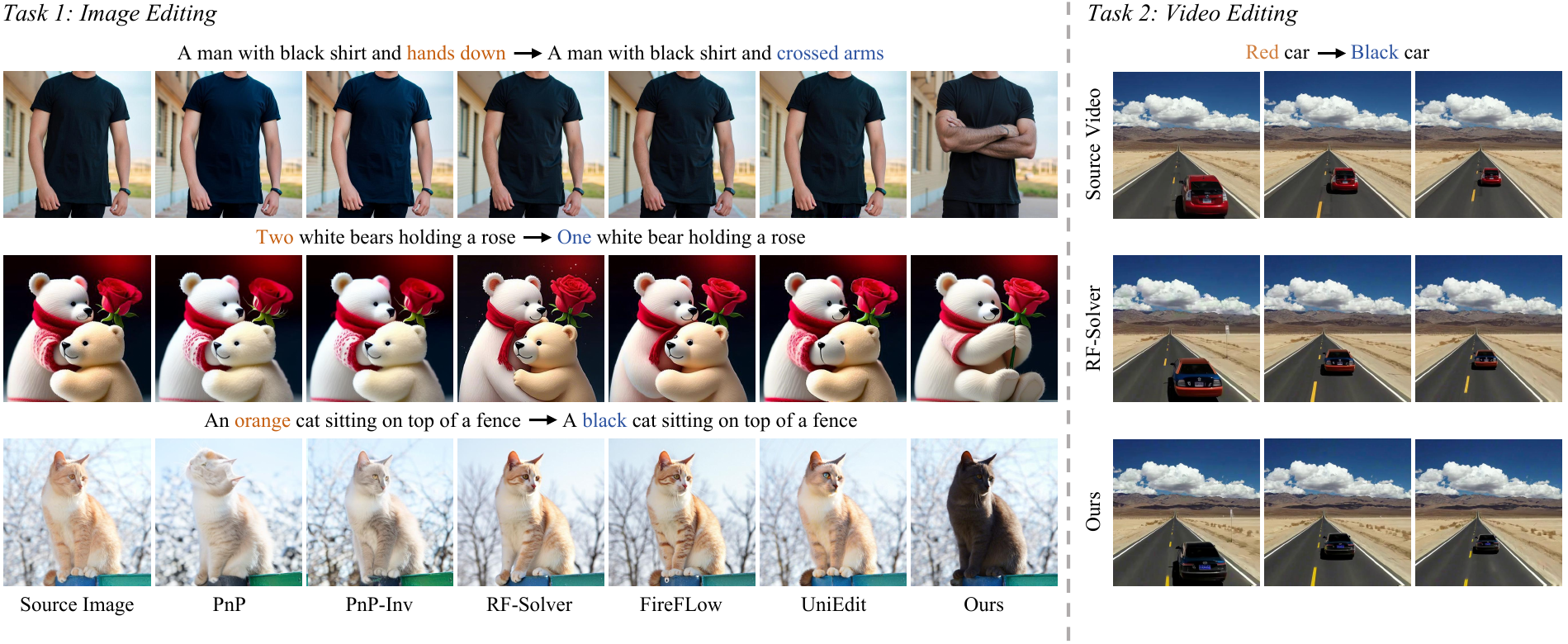}
\vspace{-0.5cm}
\captionof{figure}{
\textbf{ProEdit for image and video editing}. We propose a highly accurate, plug-and-play editing method for flow inversion that addresses the problem of excessive source image information injection, which prevents proper modification of attributes such as pose, number, and color. Our method has demonstrated impressive performance in both image editing and video editing tasks.
}
\label{first_figure} % 修改拼写错误
\vspace{-.2cm}
\end{strip}
\makeatletter
\renewcommand*{\@makefnmark}{}
\footnotetext{
  $^*$ equal contributions. \\
  \hspace*{1.8em}\textsuperscript{\Letter} corresponding authors. \hspace{5pt}\href{https://github.com/iSEE-Laboratory/ProEdit}{Code} is available
}
\begin{abstract}

Inversion-based visual editing provides an effective and training-free way to edit an image or a video based on user instructions. Existing methods typically inject source image information during the sampling process to maintain editing consistency. However, this sampling strategy overly relies on source information, which negatively affects the edits in the target image (e.g., failing to change the subject's atributes like pose, number, or color as instructed).
In this work, we propose \textbf{ProEdit} to address this issue both in the attention and the latent aspects.
In the attention aspect, we introduce KV-mix, which mixes KV features of the source and the target in the edited region, mitigating the influence of the source image on the editing region while maintaining background consistency. 
In the latent aspect, we propose Latents-Shift, which perturbs the edited region of the source latent, eliminating the influence of the inverted latent on the sampling. 
Extensive experiments on several image and video editing benchmarks demonstrate that our method achieves SOTA performance. In addition, our design is plug-and-play, which can be seamlessly integrated into existing inversion and editing methods, such as RF-Solver, FireFlow and UniEdit.

\end{abstract}    
\section{Introduction}
\label{sec:intro}

Inversion-based visual editing~\cite{meng2022sdedit,dalva2024fluxspace,miyake2025npi,mokady2023NTI,garibi2024renoise,tumanyan2023plug,ju2023direct} has emerged as a highly effective and valuable research direction, offering a powerful, training-free paradigm for modifying images and videos according to user instructions, especially the flow-based editing methods, which provides better generative abilities with fewer sampling steps. 

Most established inversion-based methods~\cite{wang2024taming,deng2024fireflow} operate by first leveraging the inverted latents from the source image as the starting point. Then, they redo the sampling process by using the target prompt to guide the sampling process towards the target image or video. To maintain fidelity to the source content, they mostly employ the source injection strategy to re-introduce source-specific information or ``nuggets" during the sampling process.
However, as shown in Fig~\ref{first_figure}, this sampling strategy overly relies on source information, no matter in the latent aspect or the attention aspect, which negatively affects the edits in the target image, especially regarding subject attributes such as color, pose, and number.

\begin{figure}
\centering
\includegraphics[width=1\linewidth]{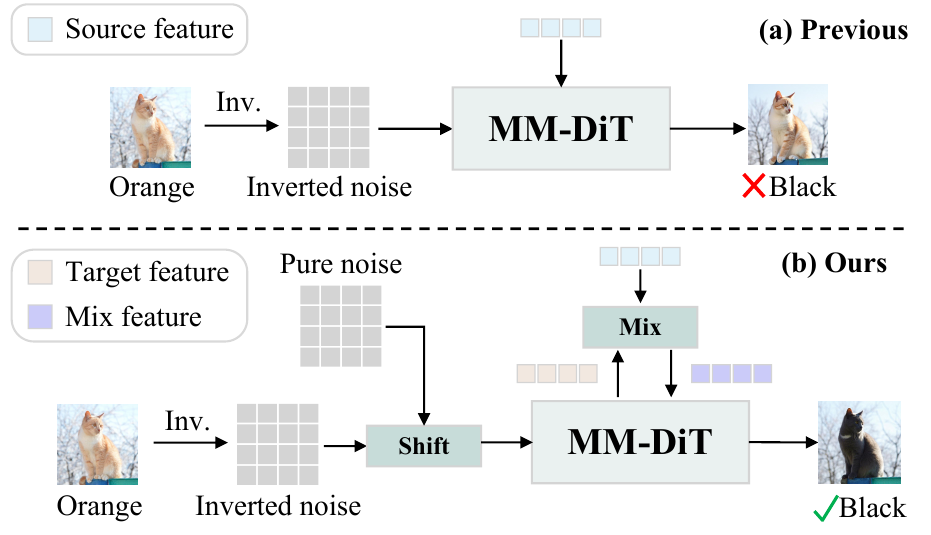}
\vskip -0.1in
\caption{Framework comparison between (a) previous methods and (b) our method. To address the issue of excessive source image information injection, we introduce the Shift module for inverted noise and the Mix module for the attention injection, alleviating the editing failures caused by these issues.
}
\vskip -0.1in
\label{fig:idea}
\end{figure}

In this work, we first conduct an in-depth investigation into the above problems and conclude that completely relying on the inverted latents for sampling and applying global attention feature injection introduces excessive source image information, leading to editing failure. Specifically, in the attention aspect, we find that the global attention feature injection strategy introduces excessive attribute-related information from the source attention, causing the model to overly focus on source information while neglecting text guidance. For the latent aspect, starting from the source image distribution creates an overly strong prior that easily leads the sampling process to reconstruct the source distribution.

Based on the above observation, we propose a novel inversion-based editing method, \textbf{ProEdit}, to eliminate the negative impact of the source image from both the attention and the latent aspects. 
For the attention aspect, we introduce \textbf{KV-mix}. We first identify the edited regions based on source and target prompts, then mix the K and V features in these regions while fully injecting source KV features in non-edited areas to preserve background consistency. This mixing mechanism applies to all attention operations without manual adjustment of heads, layers, or blocks.
For the latent aspect, we propose \textbf{Latents-Shift}. Inspired by AdaIN from style transfer, which performs structure-preserving distribution transformations, we inject random noise into the source distribution of edited regions. This reduces the influence of source image attributes while maintaining structural and background consistency.
As shown in Fig \ref{fig:idea}, our method successfully eliminates the negative effects of source image information from inversion noise and attention injection mechanisms, accurately modifying the subject's attribute while maintaining the consistency of background and non-editing content.

Through extensive experiments, we demonstrate that 1) \textbf{Effectiveness}: ProEdit can eliminate the negative impact of the source image/video on the editing content, while preserving the non-edited content, achieving state-of-the-art (SOTA) performance in various editing tasks; 2) \textbf{Plug-and-play}: Our method is plug-and-play, enabling its seamless integration into a wide range of existing inversion and editing methods; 3) \textbf{Attribute Correction}: In attribute editing, where existing methods perform poorly, our approach showcases unprecedented performance. 
Our contributions can be summarized as follows:
\begin{itemize}
    \item We investigate the issue of excessive source image information injection in inversion-based editing and identify that this problem stems from both latent initialization and attention injection mechanisms, leading to failures.
    \item We propose ProEdit, a novel training-free approach designed to solve the above problem, which can eliminate the negative impact from the source while maintaining background consistency. 
    \item Through extensive experiments, we proof that ProEdit is effective, plug-and-play and can be used in various types of image and video editing. Our code will be open-source to boost the developing of the generative community.
\end{itemize}

\section{Related Work}
\label{Related Work}

\subsection{Text-to-Visual generation}

Diffusion models have achieved significant success in the fields of text-to-visual generation, leading to the development of a series of outstanding foundational models~\cite{ho2020ddpm,song2020denoising,rombach2022high,peebles2023scalable,zhang2023controlnet}. Recently, the text-guided generation paradigm for both images and videos has been shifting from diffusion models based on U-Net~\cite{ronneberger2015u} architecture to flow models based on DiT~\cite{peebles2023scalable} architecture. Flow-based models, such as FLUX~\cite{flux2024} and HunyuanVideo~\cite{kong2024hunyuanvideo}, utilize the MMDiT~\cite{esser2024scaling} architecture and simulate a straight path between two distributions through a probability flow ordinary differential equation(ODE), enabling faster and better generation with fewer sampling steps. These T2I and T2V models also facilitate the editing of images and videos, where target images are generated based on source images and modified text prompts. 

\subsection{Text-driven Editing}
For visual editing tasks, early works focused on training-based methods~\cite{tero2021astyle,zhu2017unpair,ju2024brushnet,zheng2025panorama,li2024brushedit,brooks2023instrup2p,li2025editthinker}. These methods leverage generative models to achieve controllable image editing. As generative models have advanced, attention has shifted towards training-free editing methods, which offer greater flexibility and efficiency. Among them, inversion-based methods~\cite{wallace2023edicy} have become an important research direction for applying diffusion models to image editing tasks. DDIM inversion~\cite{song2020denoising}, as a representative method, marked a significant advancement in inversion-based image editing within diffusion models, inspiring a series of high-precision solvers~\cite{lu2025dpm,wang2024belm,zhang2024exact} aimed at minimizing inversion errors and improving sampling efficiency. Sampling-based methods introduce controlled randomness to enable more flexible editing~\cite{dong2023pnt,ju2023direct,parmar2023zeros,wu2023uncover}. On the other hand, attention-based methods achieve controllable image editing by altering the role of attention tokens~\cite{xu2023infedit,cao2023masactrl,tumanyan2023plug,tewel2025addit,li2023stylediffu,kawar2023imagic}, and these methods have gradually expanded to video editing~\cite{cai2025ditcontrol}. 

Following the trajectory of diffusion models, recent inversion methods based on flow models have mainly focused on improving inversion solvers~\cite{wang2024taming,deng2024fireflow,jiao2025uniedit,xu2025unveil} and the joint attention mechanism~\cite{xu2024headrouter,avrahami2025stable} in MM-DiT~\cite{esser2024scaling}. Although they have achieve good editing performance, these methods still overlook the negative impact of inversion strategies on the editing content. In this work, we reveal the negative impact of inversion on editing and propose ProEdit to eliminate this negative influence from both the attention and latent distribution perspectives. Notably, existing methods rely on selecting specific attention heads, layers, or block types when modifying the attention mechanism, which limits their alignment with the source image. Our method is the first to achieve this without requiring the selection of specific layers, heads, or block types.

\section{Method}

In this section, we first introduce the preliminary to understand our method in Section \ref{sec:preliminary}, then we conduct an investigation in Section \ref{sec:rethink}, analyzing the reasons why the inversion-sampling paradigm faces challenges in removing the influence of the source image on the target image's edited contents. Next, we introduce our proposed ProEdit method in Section \ref{sec:kv-mix} and \ref{sec:LS}. Our method eliminates the influence of the source image on the target image's edited contents from both the attention guidance and the initial latents in the sampling process, while maintaining the consistency of the background structure. Finally, we summarize our method's editing process in Section \ref{overall}.

\subsection{Preliminaries}
\label{sec:preliminary}
First, we introduce the preliminary knowledge to better understand our method, including the training objective of flow-based generative models and the ODE solving process. Then, we derive the inversion ODE solving process based on flow models. 

Generative models~\cite{ho2020ddpm} aim to generate data $\bm{X}_1$ that follows the real data distribution $\pi_1$ from noise $\bm{X}_0$ that follows a Gaussian distribution $\pi_0$. Recently, flow matching~\cite{albergo2023stocha,lipman2022flow,liu2022flow} has emerged as a method that learns a velocity field $\bm{v}_\theta$ to transform noise into data along a straight trajectory. The training objective is to solve the following optimization problem:
\begin{equation}
\begin{gathered}
\min_{\theta} \mathbb{E}_{\bm{Z}_0, \bm{Z}_1, t}
\left[
\left\| (\bm{Z}_1 - \bm{Z}_0) - \bm{v}_{\theta}(\bm{Z}_t, t) \right\|^2
\right], \quad \\
\bm{Z}_t = t\bm{Z}_1 + (1-t)\bm{Z}_0, \; t \in [0,1],
\end{gathered}
\label{eq:flow_matching}
\end{equation}
where $\bm{Z}_0 \sim \pi_0$ is initialized from the source distribution, and $\bm{Z}_1 \sim \pi_1$ is generated at the end of the trajectory. The term $\bm{Z}_1 - \bm{Z}_0$ represents the target velocity. The model learns a velocity field to deterministically transform random samples of Gaussian noise into target data via an Ordinary Differential Equation (ODE) defined over the continuous time interval:
\begin{equation}
    d\bm{Z}_t = \bm{v}_\theta(\bm{Z}_t, t) dt, \quad t\in[0,1]
\end{equation}
This ODE can be discretized and numerically solved by solvers as follows:
\begin{equation}
    \bm{Z}_{t_{i+1}} = \bm{Z}_{t_i} + (t_{i+1} - t_i) v_\theta(\bm{Z}_{t_i}, t_i),
\end{equation}
where $i \in \{0, \ldots, N\}$, with $t_0 = 0$ and $t_N = 1$.

Flow matching has deterministic trajectories. Its reverse process is obtained by reversing the learned flow trajectory. Starting from $\bm{Z}_1 \sim \pi_1$, the reverse ODE is given by reverse the velocity field:
\begin{equation}
    d\bm{Z}_t = -\bm{v}_\theta(\bm{Z}_t, t) dt, \quad t\in[1,0]
\end{equation}
Correspondingly, this ODE is discretized and solved using a numerical solver as follows:
\begin{equation}
    \bm{Z}_{t_{i-1}} = \bm{Z}_{t_i} - (t_{i-1} - t_i) v_\theta(\bm{Z}_{t_i}, t_i),
\end{equation}
where $i \in \{N, \ldots, 0\}$, with $t_N = 1$ and $t_0 = 0$. This inverse process generates $\bm{Z}_0 \sim \pi_0$ by utilizing the symmetry of the velocity field $v$ to ensure consistency with the forward process. Naturally, this inversion method is applied in visual reconstruction and visual editing.

\begin{figure*}[t]
\vskip -0.1in
\centering
    \includegraphics[width=0.9\linewidth]{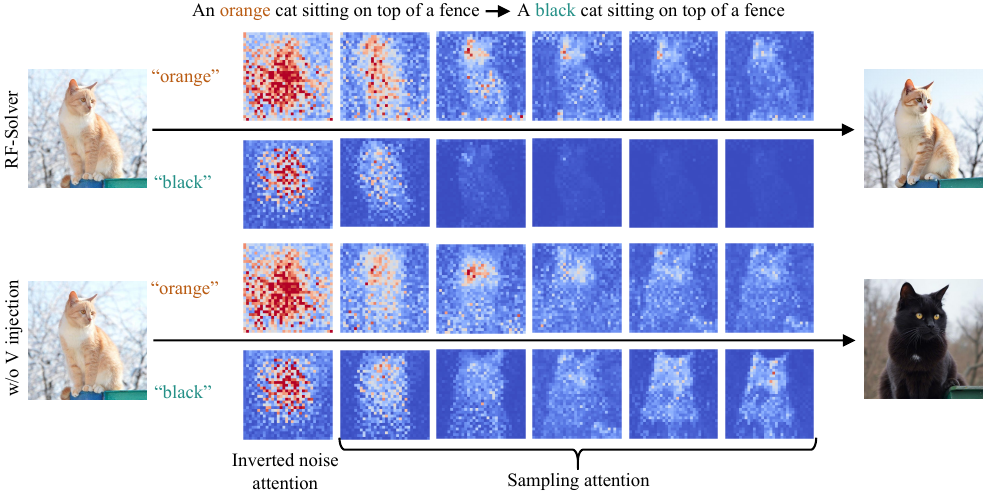}
    \caption{\textbf{Excessive source image information injection phenomenon in RF-Solver}. We validate it by visualizing the attention from source and target text tokens to the visual tokens during initial and sampling stage. In RF-Solver, the attention from the source text token to the visual tokens remains higher than that from the target text token. However, after removing attention injection, the attention from ``black'' and ``orange'' to visual tokens returns to similar levels, but some subject attributes (\eg, pose) change accordingly.}
    \label{fig:LS}
\end{figure*}

\subsection{Rethinking the Inversion-Sampling Paradigm}
\label{sec:rethink}
In this subsection, we conduct an investigation of the challenge in visual editing, and we conclude that previous works mostly rely on sampling with inverted noise and use a source attention injection mechanism to maintain background and structural consistency. This design often injects excessive source image information, leading to editing failure. The analysis is as follows.

\noindent\textbf{Attention Injection Problem.} To maintain the overall structural consistency between the target and source images, current methods~\cite{wang2024taming,deng2024fireflow} globally inject the value attention features \textit{V} into specific time steps during the sampling process, as described by the following equation:
\begin{equation}
z^{t}_{tg}(l+1)= \text{Attn}(\bm{Q}_{tg}^t,\bm{K}_{tg}^t,\bm{V}_{s}^t),
\label{eq:feature_share}
\end{equation}
where $s$ denotes the attention features corresponding to the \textbf{source prompt}, while $tg$ denotes the attention features corresponding to the \textbf{target prompt}. However, this global attention feature injection mechanism has a negative impact on the editing process. As shown in Fig~\ref{fig:LS}, after adding the attention injection mechanism, although overall consistency is enhanced, the model focus far less on prompt ``black'' than on ``orange'' in the image, representing that the subject's attributes are also forcibly injected into the target image. This issue increases the difficulty of attribute editing.

\noindent\textbf{Latent Distribution Injection Problem.} As shown in Fig~\ref{fig:LS}, the attention from ``orange'' to visual tokens is significantly higher than from ``black'', indicating that although the image is inverted back to noise, it still retains substantial source image attributes. This causes editing to fail when the gap between target and source prompts is too large.

\noindent\textbf{Summary.} The negative impact of the source image on the editing process can be attributed to two factors: global attention feature injection and the latent distribution injection. Therefore, this paper proposes ProEdit to address all these issues from corresponding aspects.

\begin{figure*}[t]
\vskip -0.1in
    \centering
    \includegraphics[width=1\linewidth]{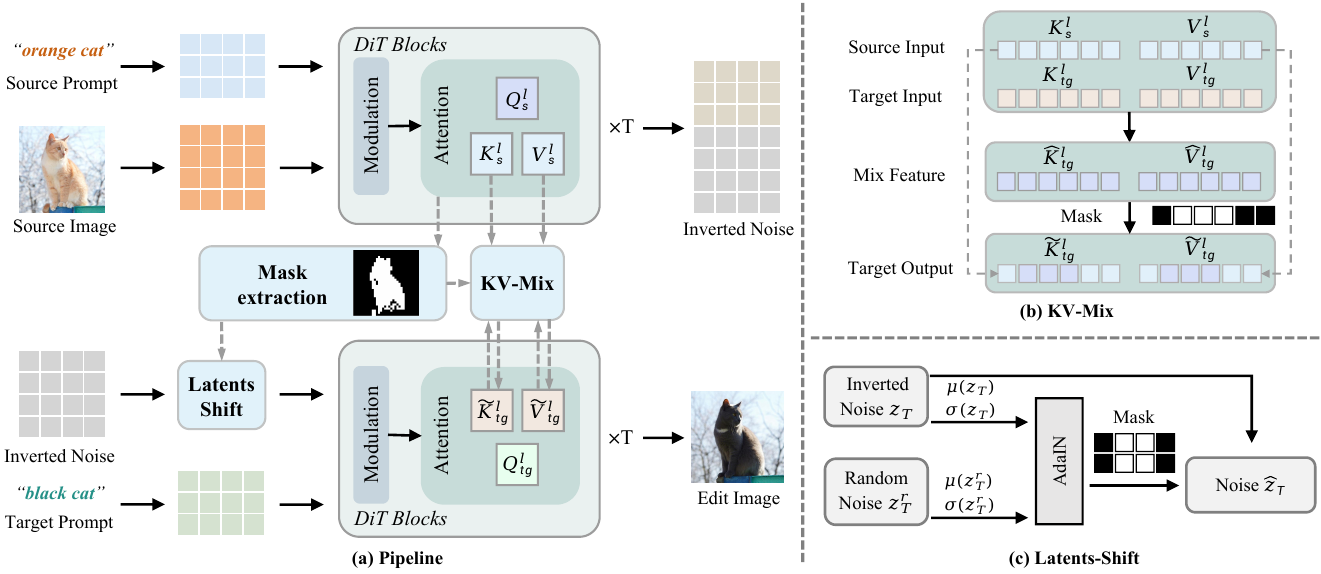}
    \vskip -0.05in
    \caption{\textbf{Pipeline of our ProEdit.} The mask extraction module identifies the edited region based on source and target prompts during the first inversion step. After obtaining the inverted noise, we apply Latents-Shift to perturb the initial distribution in the edited region, reducing source image information. In selected sampling steps, we fuse source and target attention features in the edited region while directly injecting source features in non-edited regions to achieve accurate attribute editing and background preservation simultaneously.}
    \label{fig:pipeline}
\end{figure*}

\subsection{KV-mix}
\label{sec:kv-mix}
\noindent\textbf{Motivation.} As analyzed before, previous methods use a global injection mechanism of visual and textual attention features to maintain consistency, but the excessive injection of source attention features negatively impacts the editing quality. Towards this end, our method aims to mitigate the problem with the insight that mixing source visual attention and target visual attention helps align with the target prompt while maintaining the consistency of non-edited content.

\noindent\textbf{Method.} 
Based on the above observations, we execute attention control on the visual components across all blocks, while consistently using the attention features of the target prompt for text attention to achieve effective editing guidance. To distinguish between the editing and non-editing regions, we obtain a mask $\bm{M}$ by processing the attention map to separate the editing region, for the detailed implementation please see the \textit{Supplementary File}. For the non-editing region, we apply full injection of visual attention features to maintain background consistency. For the editing region, we use a mix of source and target visual attention features to preserve the consistency of non-editing content and improve the editing quality. After extensively exploring all plausible combinations of $\bm{Q}$, $\bm{K}$, and $\bm{V}$, we found that the configuration shown in Eq.\ref{eq:KV-mix} is most conducive to achieving consistent editing. Formally, our KV-mix design is as follows:
\vspace{-0.5em} 
\begin{equation}
\begin{gathered}
\hat{{\bm{K}}}_{tg}^{l} = \delta \bm{K}_{t g}^{l}+(1-\delta) \bm{K}_{s}^{l}, \\
\hat{{\bm{V}}}_{tg}^{l} = \delta \bm{V}_{t g}^{l}+(1-\delta) \bm{V}_{s}^{l}, \\
\tilde{\bm{K}}_{t g}^{l} =
\bm{M} \odot \hat{{\bm{K}}}_{tg}^{l} +(1-\bm{M}) \odot \bm{K}_{s}^{l}, \\[4pt]
\tilde{\bm{V}}_{t g}^{l} =
\bm{M} \odot \hat{{\bm{V}}}_{tg}^{l} +(1-\bm{M}) \odot \bm{V}_{s}^{l},\\
z^{t}(l+1) =\text{Attn}\left(\bm{Q}_{t g}^{l}, \tilde{\bm{K}}_{t g}^{l}, \tilde{\bm{V}}_{t g}^{l}\right),
\end{gathered}
\label{eq:KV-mix}
\end{equation}
where $\bm{M}$ denotes the edited region that is extracted from attention map and applied only to the visual branch. To enable controllable editing strength and preservation of non-edited content, we define the \textbf{mixing strength} $\delta$ as a ratio of mix for applying attention control in the edited region, which determines the level of non-edited content preservation during editing. This attention mechanism enables precise text control for consistent editing. Since we perform the KV-mix operation only within the visual tokens, KV-mix is applied in both Double and Single Attention blocks.

\subsection{Latents-Shift}
\label{sec:LS}
\noindent\textbf{Motivation.} 
Here we aim to mitigate the problem of distribution injection while preserving the structure consistency. Inspired by AdaIN~~\cite{huang2017adain} in style transfer, which transfers color and texture distributions while preserving structural consistency, we adapt this approach to image editing.

\noindent\textbf{Method.} 
As our goal is to eliminate the influence of source image information, we directly use random noise as the style image to shift the distribution of the inverted noise. 
We improve its formula to implement the shift of the latent distribution for the editing region as follows:
\begin{equation}
    \begin{gathered}
        \tilde{z}_{T} = \sigma(z_{T}^{r}) \left( \frac{z_{T} - \mu(z_{T})}{\sigma(z_{T})} \right) + \mu(z_{T}^{r}), \\[4pt]
        \hat{z}_{T} = \bm{M} \odot(\beta \tilde{z}_{T}+(1-\beta)z_{T})+(1-\bm{M}) \odot z_{T},
    \end{gathered}
    \label{eq:Latents-shift}
\end{equation}
where $\beta$ denotes the fusion ratio between the inverted noise and pure noise, controlling the level of shift in the inverted noise distribution. $\bm{M}$ denotes the edited region, which is inherited from KV-mix to achieve the shifted inverted noise distribution for the editing region.

\subsection{Overall}
\label{overall}

The complete process of our pipeline can be summarized as Fig \ref{fig:pipeline}: During the inversion stage, the source image and source prompt are input into the model to conduct the inversion process, and $\bm{K}_{s}^{l}$ and $\bm{V}_{s}^{l}$ are cached in the fly. Then, the attention map is processed to obtain the mask of the editing region, and the inverted noise is output as the initial input for the sampling stage. 

In the sampling stage, the inverted noise first passes through the Latents-Shift module to obtain the fusion noise, which is then input into the model along with the target prompt for sampling. During the sampling process, the source visual attention features $\bm{K}_{s}^{l}$ and $\bm{V}_{s}^{l}$, obtained from the inversion, are injected through the KV-mix module, and the final model outputs the target image by multiple steps of sampling.
\section{Experiments}

\begin{table*}[t]
    \centering
    \caption{\textbf{Text-driven image editing comparison} on PIE-Bench. We report the peer-reviewed results of each baseline, and evaluate our proposed method using flow-based inversion methods RF-Solver, FireFlow, and UniEdit to demonstrate the effectiveness. The best and second-best results are shown in \textbf{bold} and \underline{underline} respectively.}
    \vspace{-2mm}
    \resizebox{2.\columnwidth}{!}{
        \begin{tabular}{ll|c|c|cc|cc|c}
        \toprule[0.12em]
        \multicolumn{2}{l|}{\multirow{2}{*}{\textbf{Method}}} & \multirow{2}{*}{\textbf{Model}} &
        \multicolumn{1}{c|}{\textbf{Structure}} &
        \multicolumn{2}{c|}{\textbf{BG Preservation}} &
        \multicolumn{2}{c|}{\textbf{CLIP Sim.$\uparrow$}} & \multirow{2}{*}{\textbf{NFE}} \\
        \multicolumn{2}{c|}{} & & 
        \textbf{Distance ($\times 10^3$)$\downarrow$} &
        \textbf{PSNR$\uparrow$} & \textbf{SSIM ($\times 10^{2}$)$\uparrow$} &
        \textbf{Whole} & \textbf{Edited} & \\
        \midrule[0.12em]
        \multicolumn{2}{l|}{P2P~\cite{hertz2022prompt}} & Diffusion & 69.43 & 17.87 & 71.14 & 25.01 & 22.44 & 100 \\
        \multicolumn{2}{l|}{PnP~\cite{tumanyan2023plug}} & Diffusion & 28.22 & 22.28 & 79.05 & 25.41 & 22.55 & 100 \\
        \multicolumn{2}{l|}{PnP-Inversion~\cite{ju2023direct}} & Diffusion & 24.29 & 22.46 & 79.68 & 25.41 & 22.62 & 100 \\
        \multicolumn{2}{l|}{EditFriendly~\cite{huberman2024edit}} & Diffusion & -- & 24.55 & 81.57 & 23.97 & 21.03 & 90 \\
        \multicolumn{2}{l|}{MasaCtrl~\cite{cao2023masactrl}} & Diffusion & 28.38 & 22.17 & 79.67 & 23.96 & 21.16 & 100 \\
        \multicolumn{2}{l|}{InfEdit~\cite{xu2023infedit}} & Diffusion & 13.78 & 28.51 & 85.66 & 25.03 & 22.22 & 72 \\
        \multicolumn{2}{l|}{RF-Inversion~\cite{rout2024semantic}} & Flow & 40.60 & 20.82 & 71.92 & 25.20 & 22.11 & 56 \\
        \midrule[0.1em]
        \multicolumn{2}{l|}{RF-Solver~\cite{wang2024taming}} & Flow & 31.10 & 22.90 & 81.90 & 26.00 & 22.88 & 60 \\
        \multicolumn{2}{l|}{\textbf{RF-Solver+Ours}} & Flow & {27.82} & {24.77} & {84.78} & {26.28} & {23.25} & {60} \\
        \midrule[0.1em]
        \multicolumn{2}{l|}{FireFlow~\cite{deng2024fireflow}} & Flow & 28.30 & 23.28 & 82.82 & 25.98 & 22.94 & 32 \\
        \multicolumn{2}{l|}{\textbf{FireFlow+Ours}} & Flow & {27.51} & {24.78} & {85.19} & {26.28} & {23.24} & {32} \\
        \midrule[0.1em]
        \multicolumn{2}{l|}{UniEdit~\cite{jiao2025uniedit}($\alpha$=0.6)} & Flow & \underline{10.14} & \underline{29.54} & \underline{90.42} & 25.80 & 22.33 & 28 \\
        \multicolumn{2}{l|}{\textbf{UniEdit($\alpha$=0.6)+Ours}} & Flow & \textbf{9.22} & \textbf{30.08} & \textbf{90.87} & {25.78} & {22.30} & {28} \\
        \midrule[0.1em]
        \multicolumn{2}{l|}{UniEdit~\cite{jiao2025uniedit}($\alpha$=0.8)} & Flow & 26.85 & 24.10 & 84.86 & \underline{26.97} & \underline{23.51} & 37 \\
        \multicolumn{2}{l|}{\textbf{UniEdit($\alpha$=0.8)+Ours}} & Flow & {24.27} & {24.82} & {85.87} & \textbf{27.08} & \textbf{23.64} & {37} \\
        \bottomrule[0.1em]
    \end{tabular}{}
    }
    \label{tab:piebench}
    \vspace{-2mm}
\end{table*}

\begin{table}[t]
    \centering
    \setlength{\tabcolsep}{3pt}
    \caption{\textbf{Quantitative comparison on Color Editing}. The best and second-best results are shown in \textbf{bold} and \underline{underline}.}
    \vspace{-2mm}
    \resizebox{\columnwidth}{!}{
        \begin{tabular}{ll|c|cc}
        \toprule[0.12em]
        \multicolumn{2}{l|}{\multirow{2}{*}{\textbf{Method}}} & 
        {\textbf{BG Preservation}} & \multicolumn{2}{c}{\textbf{CLIP Sim.$\uparrow$}} \\
        \multicolumn{2}{c|}{} & 
        \textbf{SSIM ($\times 10^{2}$)$\uparrow$} & \textbf{Whole} & \textbf{Edited} \\
        \midrule[0.12em]
        \multicolumn{2}{l|}{RF-Solver} & 80.21 & 25.61 & 20.86 \\
        \multicolumn{2}{l|}{\textbf{RF-Solver+Ours}} & \textbf{86.63} & \underline{27.30} & \textbf{22.88} \\
        \midrule[0.1em]
        \multicolumn{2}{l|}{FireFlow} & 80.14 & 26.03 & 21.02 \\
        \multicolumn{2}{l|}{\textbf{FireFlow+Ours}} & \underline{86.53} & \textbf{27.32} & 22.55 \\
        \midrule[0.1em]
        \multicolumn{2}{l|}{UniEdit} & 85.39 & 26.81 & 21.74 \\
        \multicolumn{2}{l|}{\textbf{UniEdit+Ours}} & 85.21 & 27.32 & \underline{22.56} \\
        \bottomrule[0.1em]
        \end{tabular}{}
    }
    \label{tab:coloredit}
    \vspace{-4mm}
\end{table}

\subsection{Setup}

\textbf{Baseline.} We mainly compare our methods with previous state-of-the-art turning-free visual editing methods. For text-driven image editing, we compare our ProEdit with diffusion-based methods: P2P~\cite{hertz2022prompt}, PnP~\cite{tumanyan2023plug}, PnP-Inversion~\cite{ju2023direct}, EditFriendly~\cite{huberman2024edit}, MasaCtrl~\cite{cao2023masactrl}, and InfEdit~\cite{xu2023infedit}, along with flow-based methods: RF-Inversion~\cite{rout2024semantic}, RF-Solver~\cite{wang2024taming}, Fire-Flow~\cite{deng2024fireflow}, and UniEdit~\cite{jiao2025uniedit}. For text-driven video editing, we compare our ProEdit with FateZero~\cite{qi2023fatezero}, Flatten~\cite{cong2023flatten}, Tokenflow~\cite{geyer2023tokenflow}, and RF-Solver~\cite{wang2024taming}. \\
\textbf{Datasets.} For text-driven image editing, we evaluate our method based on the PIE-Bench~\cite{ju2023direct}, which contains 700 images with 10 different editing types. For text-driven video editing, we collected 55 text-video editing pairs with a resolution of $480 \times 480$, $540 \times 960$ or $960 \times 540$, consist of $40$ to $120$ frames, including the videos sourced from DAVIS dataset~\cite{pont20172017} and online platforms. The prompts are derived from ChatGPT or contributed by the authors.  \\
\textbf{Metrics.} For text-driven image editing, to evaluate edit-irrelevant context preservation, we use structure distance~\cite{tumanyan2022splic}, PSNR~\cite{huynh2008scope} and SSIM~\cite{wang2004image} for annotated unedited regions. The performance of the edits is assessed using CLIP similarity~\cite{radford2021clip} for both the whole image and the edited regions. For text-driven video editing, we follow the metrics proposed in VBench~\cite{huang2024vbench,zheng2025vbench2}, including Subject Consistency, Motion Smoothness, Aesthetic Quality, and Imaging Quality. \\ 
\textbf{Implementation.} We primarily product experiments using FLUX.1-[dev]~\cite{flux2024} for image editing and HunyuanVideo-720p~\cite{kong2024hunyuanvideo} for video editing. For image editing, we have made ProEdit plug-and-play for flow-based inversion methods: RF-Solver, FireFlow, and UniEdit. Notably, UniEdit uses $\alpha$ to denote the delay injection rate, and experiments were conducted with $\alpha=0.6$ and $\alpha=0.8$. In our experiments, unless otherwise specified, the delay rate $\alpha$ in UniEdit is set to 0.8. We set the sampling step for image editing to 15. For video editing, we have made ProEdit plug-and-play for RF-Solver. We set the sampling step for video editing to 25.

\subsection{Text-driven Image Editing}
\textbf{Quantitative Evaluation.} Table \ref{tab:piebench} presents the quantitative results for text-driven image editing.
The results in the table show that with our ProEdit, flow-based inversion methods achieved superior results in image editing. Notably, ProEdit with the UniEdit inversion method achieves \textbf{state-of-the-art} performance in both source content preservation and editing quality. 

\begin{figure*}
\vskip -0.1in
    \centering
    \includegraphics[width=1\linewidth]{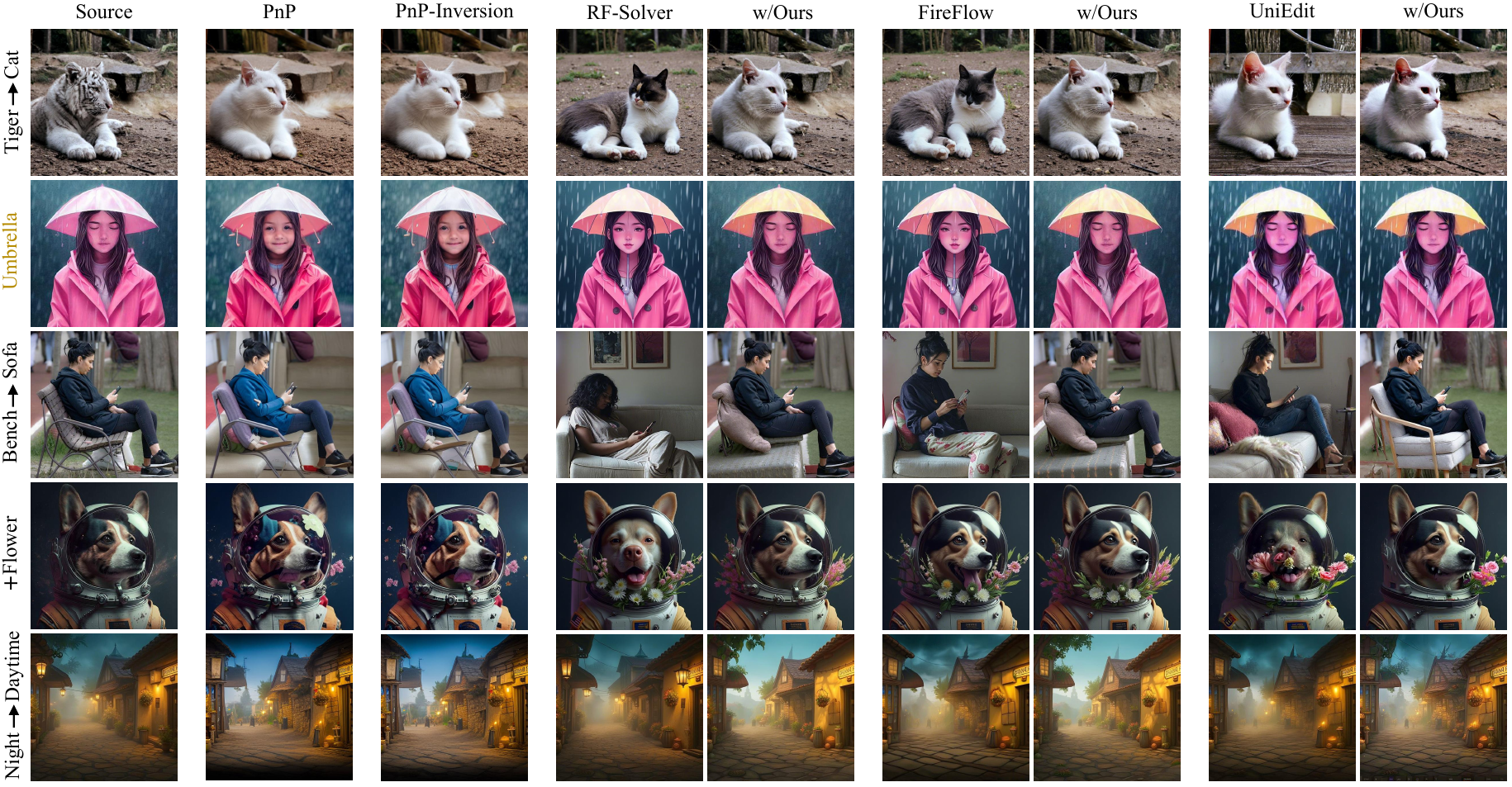}
    \vspace{-6mm}
    \caption{\textbf{Qualitative comparison on image editing}. With our method, various flow-based inversion methods achieve more appropriate editing while preserving the consistency of background and non-editing content.}
    \label{fig:imageedit}
\end{figure*}

\noindent\textbf{Qualitative Evaluation.} We compare the performance of our method with several baselines across different types of editing requirements in Fig \ref{fig:imageedit}. The baseline methods often fail to maintain the consistency of non-editing attributes such as background and posture, or fail to achieve satisfactory editing results. In contrast, our method achieves high-quality editing results while maintaining the consistency of non-editing content.

\begin{figure*}[t]
\vskip -0.1in
    \centering
    \includegraphics[width=1\linewidth]{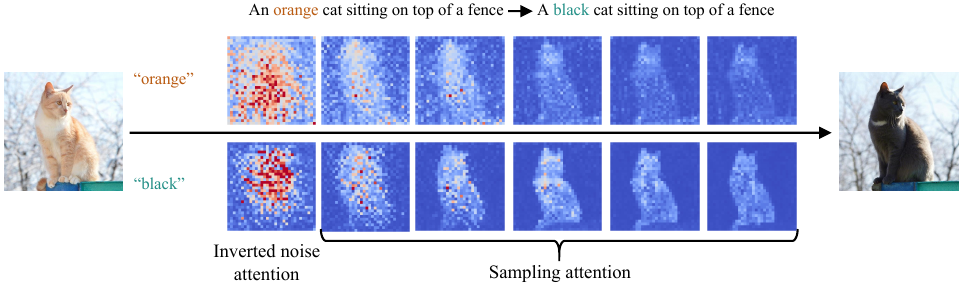}
    \vspace{-6mm}
    \caption{\textbf{Visualization of attention map after performing ProEdit}. The initial distribution is shifted to target prompt and during the sampling, the model can accurately edit the image while maintaining non-editing attribute and background concsistent.}
    \label{fig:attn_experiment}
    \vspace{-4mm}
\end{figure*}

\noindent\textbf{Color Editing.} To validate that our method addresses the ``latent distribution injection'' issue overlooked by previous methods, We conducted experiments on the color editing task in PIE-Bench, which is significantly affected by the latent distribution. Table \ref{tab:coloredit} shows the quantitative results for color editing. With our ProEdit, all flow-based inversion methods achieved impressive results. This supports the motivation behind our proposed Latents-Shift module. The AdaIN-based Latents-Shift helps the editing process break free from the constraints imposed by the source image distribution. We further validate it by visualizing the attention map after adding Latents-Shift in Fig~\ref{fig:attn_experiment}.

\subsection{Text-driven Video Editing}
\textbf{Quantitative Evaluation.} Table \ref{tab:videoedit} presents the quantitative results for text-driven video editing. For each metric, we report the average score of all videos. The results in the table show that with our ProEdit, flow-based inversion methods achieved superior results in video editing. This proves the versatility of our method for flow-based models, demonstrating its applicability to video editing tasks and its ability to improve editing performance.

\begin{figure}
    \centering
    \vspace{-1mm}
    \includegraphics[width=1\linewidth]{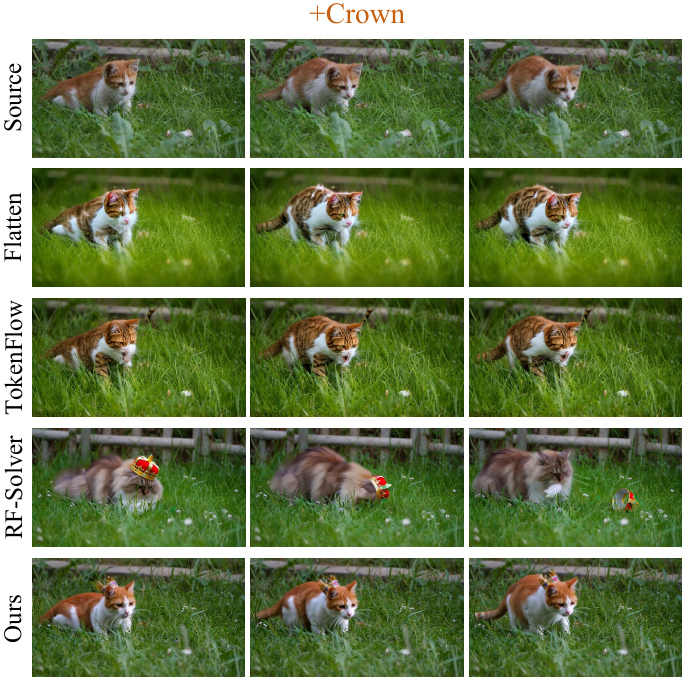}
    \caption{\textbf{Qualitative comparison on video editing}. The video comprises 48 frames with a resolution of $540 \times 960$.}
    \label{fig:videoedit}
    \vspace{-2mm}
\end{figure}

\noindent\textbf{Qualitative Evaluation.} We compare the performance of our method with several methods in Fig \ref{fig:videoedit}. The baseline methods often fail to maintain the consistency of non-editing attributes such as background and posture, or fail to achieve satisfactory editing results. In contrast, ProEdit achieves high-quality editing while maintaining spatial and temporal consistency.

\subsection{Ablation Study}
We conduct ablation study on PIE-Bench. First, we evaluate the effectiveness of each module we proposed and validate their synergistic effect. Then, we explore the best combinations within the attention mixing injecting mechanism.

\begin{table}[t]
    \centering
    \caption{\textbf{Text-driven video editing comparison}. We report the peer-reviewed results of each baseline, and evaluate our proposed method using flow-based inversion method RF-Solver. The best and second-best results are shown in \textbf{bold} and \underline{underline}.}
    \vspace{-2mm}
    \resizebox{\columnwidth}{!}{
        \begin{tabular}{ll|c|c|c|c}
        \toprule[0.12em]
        \multicolumn{2}{l|}{\textbf{Method}} & \textbf{SC} $\uparrow$ & \textbf{MS} $\uparrow$ & \textbf{AQ} $\uparrow$ & \textbf{IQ} $\uparrow$ \\
        \midrule[0.12em]
        \multicolumn{2}{l|}{FateZero~\cite{qi2023fatezero}} & 0.9612 & 0.9740 & 0.6004 & 0.6556 \\
        \multicolumn{2}{l|}{Flatten~\cite{cong2023flatten}} & 0.9690 & 0.9830 & 0.6318 & 0.6678 \\
        \multicolumn{2}{l|}{TokenFlow~\cite{geyer2023tokenflow}} & 0.9697 & 0.9897 & 0.6436 & 0.6817 \\
        \multicolumn{2}{l|}{RF-Solver~\cite{wang2024taming}} & \underline{0.9708} & \underline{0.9906} & \underline{0.6497} & \underline{0.6866} \\
        \multicolumn{2}{l|}{\textbf{RF-Solver+Ours}} & \textbf{0.9712} & \textbf{0.9920} & \textbf{0.6518} & \textbf{0.6936} \\
        \bottomrule[0.1em]
        \end{tabular}{}
    }
    \label{tab:videoedit}
    \vspace{-4mm}
\end{table}

\begin{table}[t]
    \centering
    \small  
    \renewcommand{\arraystretch}{1}
    \caption{\textbf{Quantitative comparison for the ablation study}. KV-m, LS mean KV-mix, Latents-Shift in our method. The best and second-best results are shown in \textbf{bold} and \underline{underline} respectively.}
    \vspace{-2mm}
    \resizebox{0.85\columnwidth}{!}{  % 从\columnwidth改为0.95\columnwidth
        \begin{tabular}{ll|cc|cc}
        \toprule[0.12em]
        \multicolumn{2}{l|}{\multirow{2}{*}{\textbf{Method}}} & \multirow{2}{*}{\textbf{KV-m}} & \multirow{2}{*}{\textbf{LS}} & \multicolumn{2}{c}{\textbf{CLIP Sim.$\uparrow$}} \\
        \multicolumn{2}{c|}{} & & & \textbf{Whole} & \textbf{Edited} \\
        \midrule[0.12em]
        \multicolumn{2}{l|}{} & & & 26.00 & 22.88 \\
        \multicolumn{2}{l|}{\textbf{RF-Solver}} & {\checkmark} & & \underline{26.21} & \underline{23.21} \\
        \multicolumn{2}{l|}{} & \checkmark & \checkmark & \textbf{26.28} & \textbf{23.25} \\
        \midrule[0.1em]
        \multicolumn{2}{l|}{} & & & 25.98 & 22.94 \\
        \multicolumn{2}{l|}{\textbf{FireFlow}} & {\checkmark} & & \underline{26.22} & \underline{23.18} \\
        \multicolumn{2}{l|}{} & \checkmark & \checkmark & \textbf{26.28} & \textbf{23.24} \\
        \midrule[0.1em]
        \multicolumn{2}{l|}{} & & & 26.97 & 23.51 \\
        \multicolumn{2}{l|}{\textbf{UniEdit}} & {\checkmark} & & \underline{27.02} & \underline{23.54} \\
        \multicolumn{2}{l|}{} & \checkmark & \checkmark & \textbf{27.08} & \textbf{23.64} \\
        \bottomrule[0.1em]
        \end{tabular}{}
    }
    \label{tab:ablation}
    \vspace{-4mm}
\end{table}

\noindent\textbf{The Synergistic Effect Analysis.} We evaluate the effectiveness of the proposed \textbf{KV-mix} and \textbf{Latents-Shift} in Table \ref{tab:ablation}. Note that without the modules we proposed, each method is evaluated using its original source code setup. When the KV-mix feature injection mechanism is applied to replace the original feature injection mechanism, we observe a significant improvement in CLIP similarity due to the reduced influence of source features in the attention. After incorporating the Latents-Shift module, the CLIP similarity is further enhanced as the influence of the source image on the inversion noise latent distribution is eliminated. In summary, the various modules of ProEdit work synergistically to improve the editing results.

\noindent\textbf{The Attention Feature Combination Effect Analysis.} We evaluated the effectiveness of different attention feature combinations applied in the fusion injection mechanism using the RF-Solver inversion method on PIE-Bench to verify the superiority of our proposed \textbf{KV-mix} module. The quantitative results for different attention feature combinations is shown in \textit{Supplementary File}. Note that the \bm{$V$} attention feature is the most important attention feature for editing quality, so all the attention feature combinations we evaluated include V. Among the four combinations we evaluate, the \textbf{KV} combination achieved the best performance in both background consistency preservation and editing quality. Therefore, we adopted the KV fusion injection mechanism and designed the KV-mix module.

\section{Conclusion}
In this work, we identified the issue of excessive injection of source image information caused by the "inverted latent with global injection" strategy used in existing flow-based inversion editing methods, which leads to sacrificing editing quality in order to maintain background consistency with the source image during the editing process. We introduce ProEdit, a novel, training-free method that addresses this issue by proposing the KV-mix and Latents-Shift modules from both the attention and latent perspectives, aiming to eliminate the negative impact of excessive source image information injection on editing quality. Extensive experiments show that ProEdit can be seamlessly integrated into existing flow-based inversion methods, achieving high background consistency and excellent editing quality simultaneously.

{
    \small
    \bibliographystyle{ieeenat_fullname}
    \bibliography{main}
}

\clearpage
\setcounter{page}{1}
\maketitlesupplementary

\appendix

\section{Extracting Mask From Attention Map}
\begin{figure}
    \centering
    \includegraphics[width=1\linewidth]{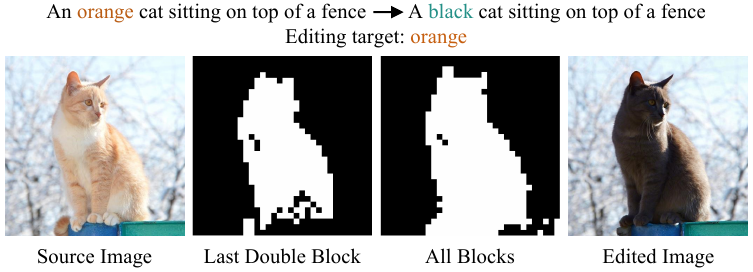}
    \caption{\textbf{A visual comparison of the editing region mask extracted from the last Double block and all blocks.} Using "orange" as the editing target, the editing region masks extracted from both the last Double block and all blocks effectively segment the editing region.}
    \label{fig:mask}
\end{figure}

When extracting the mask from the attention map, we adopted a strategy different from DitCtrl\cite{cai2025ditcontrol}. In our observation, the attention map of the last Double block effectively associates the relevant regions of text and image. Fig \ref{fig:mask} presents an example showing that both the editing region mask extracted from the last Double block's attention map and the mask extracted from the average of all attention maps successfully segment the editing and non-editing regions. Additionally, this mask extraction method reduces memory consumption. Therefore, we extract our editing region mask $\bm{M}$ from the attention map of the last Double block. Notably, we always extract the mask from the first step of inversion or the last step of sampling, as the images at these time steps are least affected by noise and have the best text-to-image correlation.

Due to the downsampling operation performed in the feature space relative to the pixel space, the extracted mask has a coarser granularity and cannot fully cover the boundary regions of the editing target in the pixel space. Therefore, we apply a diffusion operation to the mask, expanding it outward by one step to obtain a coarser mask that can fully cover the editing area. Given the strong global adaptability of generative models, this relatively coarse masking is sufficient for semantic alignment. The boundary coverage between the editing and non-editing regions helps smooth the edges and avoid image artifacts. The target object of the mask extraction can be identified by the noun of the editing object or through an externally provided mask for more flexible control.

\section{Implementation Details}
For image and video editing, we set the mix strength $\delta=0.9$ to balance source content preservation and editing performance. The fusion ratio $\beta$ is set to 0.25 to achieve the best editing results. At each timestep, the feature fusion injection mechanism is applied to all Double and Single blocks. We fine-tune the hyperparameters of the attention feature fusion injection steps to obtain better image and video editing results. We use the official implementations of all baseline methods and adjust their hyperparameters to achieve satisfactory performance.

\section{Quantitative Results of Attention Feature Combination Effect}
\begin{table}[t]
    \centering
    \caption{\textbf{Quantitative experiments on different attention feature combinations}. The best and second-best results are shown in \textbf{bold} and \underline{underline} respectively.}
    \resizebox{\columnwidth}{!}{
        \begin{tabular}{ll|cc|cc}
        \toprule[0.12em]
        \multicolumn{2}{l|}{\multirow{2}{*}{\textbf{Method}}} & \multicolumn{2}{c|}{\textbf{BG Preservation}} & \multicolumn{2}{c}{\textbf{CLIP Sim.$\uparrow$}} \\
        \multicolumn{2}{c|}{} & \textbf{PSNR$\uparrow$} & \textbf{SSIM ($\times 10^{2}$)$\uparrow$} & \textbf{Whole} & \textbf{Edited} \\
        \midrule[0.12em]
        \multicolumn{2}{l|}{\bm{$Q \& V$}} & 24.04 & 82.24 & 26.16 & 23.04 \\
        \multicolumn{2}{l|}{\bm{$Q \& K \& V$}} & 24.51 & 83.04 & 26.20 & 22.97 \\
        \multicolumn{2}{l|}{\bm{$V$}} & 23.69 & 81.68 & 26.26 & 23.15 \\
        \multicolumn{2}{l|}{\bm{$K \& V$}} & \textbf{24.77} & \textbf{84.78} & \textbf{26.28} & \textbf{23.25} \\
        \bottomrule[0.1em]
        \end{tabular}{}
    }
    \label{tab:attention ablation}
\end{table}

Table \ref{tab:attention ablation} shows the quantitative results of different attention feature combinations in the fusion injection mechanism. Among the four combinations evaluated, the \textbf{KV} combination demonstrated satisfactory results in both background consistency and editing quality. Therefore, we adopted the KV fusion injection mechanism and designed the KV-mix module.

\section{More Qualitative Results for Image Editing}
\begin{figure*}[t]
    \centering
    \includegraphics[width=1\linewidth]{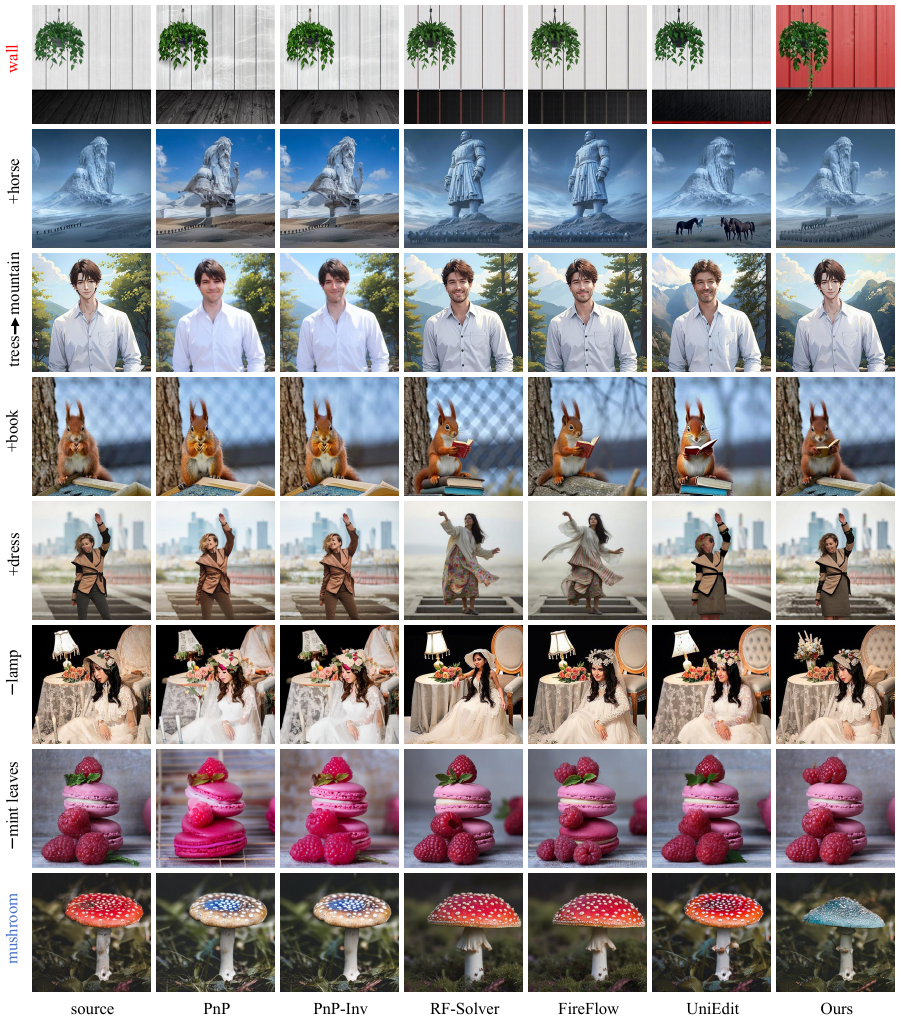}
    \caption{\textbf{More qualitative comparison of image editing on PIE-Bench\cite{ju2023direct}.}}
    \label{fig:more_image}
\end{figure*}

Here we provide more qualitative results for image editing in Fig \ref{fig:more_image}. In cases where other inversion-based editing methods\cite{ju2023direct,tumanyan2023plug,wang2024taming,jiao2025uniedit,deng2024fireflow} fail, result in insufficient edits, or fail to maintain consistency, our method successfully achieves semantically consistent editing and demonstrates impressive performance. It is worth noting that in our qualitative results (\textbf{3rd, 5th, and 6th rows in Fig \ref{fig:more_image}}), our method is able to effectively preserve human characteristics when editing human-centric images.

\section{More Qualitative Results for Video Editing}

\begin{figure*}[t]
    \includegraphics[width=1\linewidth]{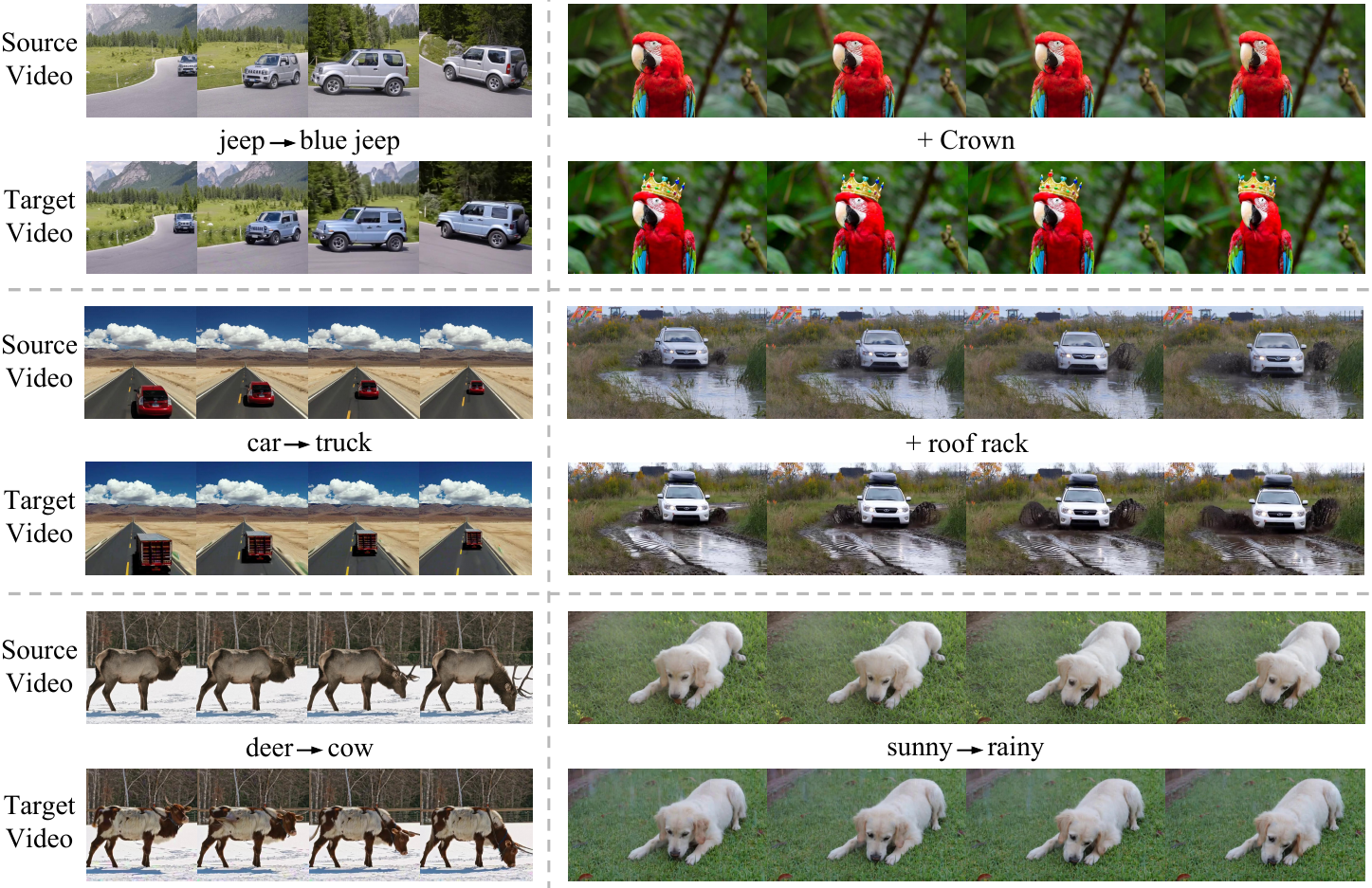}
    \caption{\textbf{More video editing results.}}
    \label{fig:more_video}
\end{figure*}

Here we provide more qualitative results for video editing in Fig \ref{fig:more_video}. Our method demonstrates impressive performance across a wide range of video editing tasks, while maintaining temporal consistency and preserving the original motion patterns.

\section{Editing by Instruction}
\begin{figure*}[b]
    \includegraphics[width=1\linewidth]{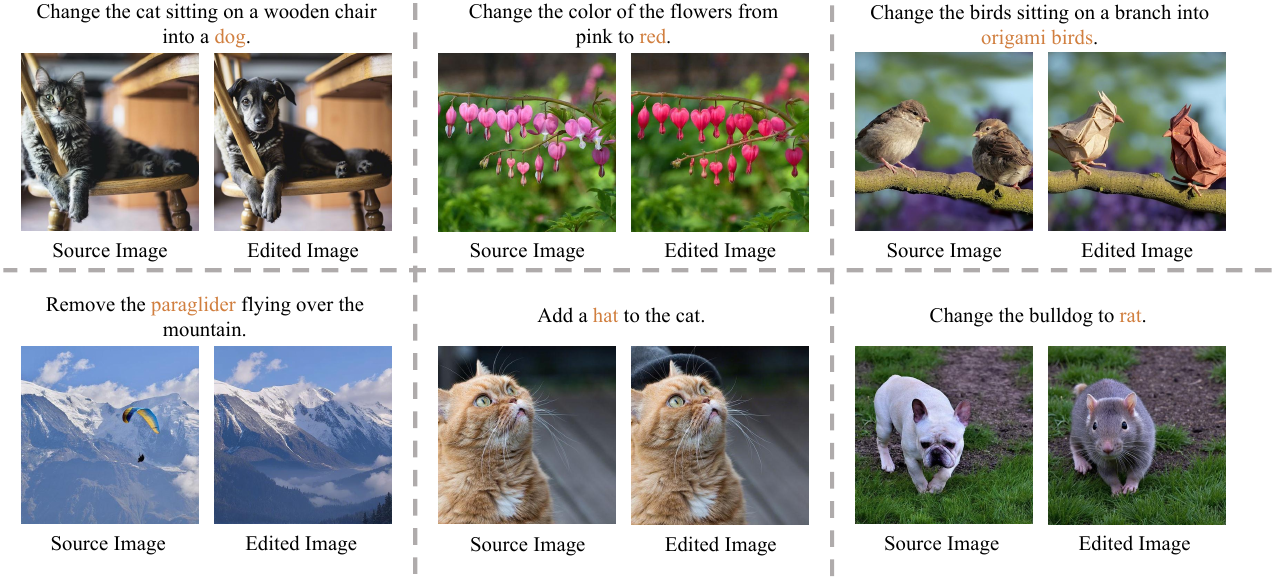}
    \caption{\textbf{Qualitative results of image editing based on editing instruction.} The actual input editing instruction are shown above each source image and its corresponding edited image.}
    \label{fig:editing_instruction}
\end{figure*}

To lower the barrier for using our method and make it more user-friendly, we introduce a large language model Qwen3-8B\cite{qwen3technicalreport} to enable editing based on editing instructions. Fig \ref{fig:editing_instruction} shows the qualitative results of our method based on editing instruction. With the assistance of a large language model, our method can directly perform edits guided by editing instructions.

\end{document}